\documentclass{article} 
\usepackage{lets_keepit_simple,times}

\usepackage{subcaption}     
\usepackage{hyperref}       
\usepackage{url}            
\usepackage{booktabs}       
\usepackage{amsfonts}       
\usepackage{nicefrac}       
\usepackage{microtype}      
\usepackage{graphicx}

\usepackage{tabulary}
\usepackage{float}

\title{\center{Let's keep it simple,}\\
 Using simple architectures to outperform deeper and more complex architectures 
}

\author{Seyyed Hossein Hasanpour\textsuperscript{1} \thanks{Corresponding author \newline \textsuperscript{1} Current affiliation: Arvenware Ltd \newline \textsuperscript{2} Current affiliation: DecorAr \newline \textsuperscript{3} Current affiliation: Microsoft} \\
Department of Computer Science\\
Islamic Azad University, Science and Research branch\\
Tehran, Iran\\
\texttt{St.h.hasanpour@iauamol.ac.ir} \\
\And
Mohammad Rouhani\textsuperscript{2} \\
Computer Vision Researcher, Technicolor R\&I\\
Rennes, France\\
\texttt{Mohammad.Rouhani@technicolor.com} \\
\And
Mohsen Fayyaz\textsuperscript{3} \\
Deep Learning Researcher, Sensifai \hspace{1cm}\hspace{1cm}\hspace{1cm}\\
Belgium\\
\texttt{Fayyaz@sensifai.com} \\
\And
Mohammad Sabokrou \\
Institute for Research in Fundamental Sciences (IPM)\\
Tehran, Iran\\
\texttt{Sabokro@ipm.ir}
}

\begin{document}

\maketitle

\begin{abstract}
Major winning Convolutional Neural Networks (CNNs), such as AlexNet, VGGNet, ResNet, GoogleNet, include tens to hundreds of millions of parameters, which impose considerable computation and memory overhead. This limits their practical use for training, optimization and memory efficiency. On the contrary, light-weight architectures, being proposed to address this issue, mainly suffer from low accuracy. These inefficiencies mostly stem from following an ad hoc procedure. We propose a simple architecture, called SimpleNet, based on a set of designing principles, with which we empirically show, a well-crafted yet simple and reasonably deep architecture can perform on par with deeper and more complex architectures. SimpleNet provides a good tradeoff between the computation/memory efficiency and the accuracy. Our simple 13-layer architecture outperforms most of the deeper and complex architectures to date such as VGGNet, ResNet, and GoogleNet on several well-known benchmarks while having 2 to 25 times fewer number of parameters and operations. This makes it very handy for embedded systems or systems with computational and memory limitations. We achieved state-of-the-art result on CIFAR10 outperforming several heavier architectures, near state of the art on MNIST and highly competitive results on CIFAR100 and SVHN. We also outperformed the much larger and deeper architectures such as VGGNet and popular variants of ResNets among others on the ImageNet dataset. Models are made available at: \href{url}{https://github.com/Coderx7/SimpleNet} 
\end{abstract}

\section{Introduction}
Since the resurgence of neural networks, deep learning methods have been gaining huge success in diverse fields of applications, amongst which, semantic segmentation, classification, object detection, image annotation and natural language processing are few to mention \cite{Guo_deep_learning_review_2015}.
What has made this enormous success possible is the ability of deep architectures to do feature learning automatically, eliminating the need for a feature engineering stage. In this stage which is the most important one amongst others, the preprocessing pipelines and data transformation are designed using human ingenuity and prior knowledge \cite{bengio_representation_2013} and has a profound effect on the end result. It is highly dependent on the level of engineers experience and expertise and if done poorly the result would be disappointing. It however, cannot scale or be generalized for other tasks well. Furthermore, in deep learning methods, instead of manual and troublesome feature engineering,  feature learning is carried out automatically in an efficient way. Deep methods also scale very well to different tasks of different essence. This proved extremely successful which one can say by looking at the diverse fields it has been being used.\\
CNNs, have been one of the most popular deep learning methods and also a major winner in many computer vision and natural language processing related tasks lately \cite{Simonyan_VGG_2014,Szegedy_googlenet_2015,He_ResNet_2015}. Since CNNs take into account the locality of the input, they can find different levels of correlation through a hierarchy of consecutive application of convolution filters. This way they are able to find and exploit different levels of abstractions in the input data and using this perform very well on both coarse and fine level details. Therefore the depth of a CNN plays an important role in the discriminability power the network offers. The deeper the better.\\
What all of the recent architectures have in common is the increasing depth and complexity of the network that provides better accuracy for the aforementioned tasks. The winner of the ImageNet Large Scale Visual Recognition Competition 2015 (ILSVRC) \cite{Russakovsky_ImageNet_2015} has achieved its success using a very deep architecture of 152 layers \cite{He_ResNet_2015}. The runner up also deploys a deep architecture of 22 layers \cite{Szegedy_googlenet_2015}. This trend has proved useful in the natural language processing benchmarks as well \cite{Sercu_VeryDeepMulti_2015}.\\
While this approach has been useful, there are some inevitable issues that arise when the network gets more complex. Computation and memory usage cost and overhead is one of the critical issues that is caused by the excessive effort put on making networks deeper and more complex in order to make them perform better. This has a negative effect on the expansion of methods and applications utilizing deep architectures. Despite the existence of various techniques for improving the learning algorithm, such as different initialization algorithms \cite{Glorot_Xavier_Understanding_difficulty_training_dnn_2010, He_PReLU_2015, Hinton_Distillation_2015, Mishkin_AllYouNeedIsGoodInit_2015, Saxe_ExactSolution_2013}, normalization and regularization method and techniques \cite{Graham_FractionalMaxpooling_2014, Goodfellow_MaxoutNetwork_2013, Ioffe_BatchNorm_incepv2_2015, Wager_dropout_training_2013, Wan_Regularization_Using_DropConnect_2013}, non-linearities \cite{Clevert_Fast_n_accurat_ELU_2015, He_PReLU_2015, Maas_RectifierNonlinearities_2013, Nair_ReLU_RBM_2010} and data-augmentation tricks \cite{AlexKrizhevsky_imgnet_2012, Graham_FractionalMaxpooling_2014, Simonyan_VGG_2014, Wu_deepImage_2015, Xu_EmpiricalEvalRectified_2015}, they are most beneficial when used on an already well performing architecture. In addition, some of these techniques may even impose more computational and memory usage overhead \cite{Goodfellow_MaxoutNetwork_2013, Ioffe_BatchNorm_incepv2_2015}. Therefore, it would be highly desirable to propose efficient architectures with smaller number of layers and parameters that are as good as their deeper versions. Such architectures can then be further tweaked using novel advancements in the literature.\\ 
The main contribution of our work is the proposal of a simple architecture, with minimum reliance on new features that outperforms almost all deeper architectures with 2 to 25 times fewer parameters. Our architecture, SimpleNet, can be a very good candidate for many scenarios, especially for deploying in the embedded devices. It can be further compressed using methods such as DeepCompression \cite{Han_deep_compression_2015} and thus its memory consumption can be decreased drastically.\\
We intentionally imposed some limitation on ourselves when designing the architecture and tried to create a mother architecture with minimum reliance on new features proposed recently, to show the effectiveness of a well-crafted yet simple convolutional architecture. It is clear when the model performs well in spite of all limitations, relaxing those limitations can further boost the performance with little to no effort which is very desirable. This performance boost however has direct correlation with how well an architecture is designed. However a fundamentally badly designed architecture would not be able to harness the advantages because of its inherent [flawed] design,  therefore we also provide the intuitions behind the overall design choices.\\

The rest of the paper is organized as follows: Section \ref{sec:related} presents the most relevant works. In Section \ref{sec:arch} we present our architecture and the set of designing principles used in the design of the architecture. In Section \ref{sec:exp} the experimental results are presented conducted on 5 major datasets (CIFAR10, CIFAR100, SVHN, MNIST and ImageNet) and more details about the architecture and different changes pertaining to each dataset are explained. Finally, conclusions and future work are summarized in Section \ref{sec:conclusion} and acknowledgment is covered in section \ref{sec:ackn}.
\section{Related Works} \label{sec:related}
In this section, we review the latest trends in related works in the literature. We categorize them into 4 sections and explain them briefly. 
\subsection{Complex networks}
Designing more effective networks was desirable and attempted from the advent of neural networks \cite{Fukushima_Neocognitron_Beginning_1979, Fukushima_Neocognitron_Self_Orgenizing_NN_1980, Ivankhnenko_Polynomial_theory_1971}. With the advent of deep learning methods, this desire manifested itself in the form of creating deeper and more complex architectures \cite{Ciresan_Deep_big_simple_nn_2010, Ciresan_A_committee_of_nn_2011, Ciresan_Multi_colmn_dnn_traffic_sign_2012, He_ResNet_2015, AlexKrizhevsky_imgnet_2012, Simonyan_VGG_2014, Srivastava_HighwayNets_2015, Szegedy_googlenet_2015, Zagoruyko_WRN_2016}. This was first attempted and popularized by \cite{Ciresan_Deep_big_simple_nn_2010} training a 9 layer MLP on GPU which was then practiced by other researchers \cite{ Ciresan_A_committee_of_nn_2011, Ciresan_Multi_colmn_dnn_traffic_sign_2012, Ciregan_Multi_column_dnn_img_cls_2012, He_ResNet_2015, AlexKrizhevsky_imgnet_2012, Simonyan_VGG_2014, Srivastava_HighwayNets_2015, Szegedy_googlenet_2015, Zagoruyko_WRN_2016}.\\
In 2012 \cite{AlexKrizhevsky_imgnet_2012} created a deeper version of LeNet5 \cite{Lecun_GradientBased_CNN_1998} with 8 layers called AlexNet, unlike LeNet5, It had local contrast normalization, ReLU \cite{Nair_ReLU_RBM_2010} nonlinearity instead of Tanh, and a new regularization layer called Dropout \cite{Hinton_preventingCoAdapt_2012}, this architecture achieved state of the art on ILSVRC 2012. The same year, \cite{Le_Building_HighlevelFeats_2013} trained a gigantic network with 1 billion parameters, their work was later proceeded by \cite{Coates_deepLearning_COTS_HPC_2013} which an 11 billion parameter network was trained. Both of them were ousted by much smaller network AlexNet  \cite{AlexKrizhevsky_imgnet_2012}.\\
In 2013 \cite{Lin_NIN_2013} released their 12 layer, NIN architecture, they built micro neural networks into convolutional neural network using $1 \times 1$ kernels. They also used global pooling instead of fully connected layers at the end acting as a structural regularizer that explicitly enforces feature maps to be confidence maps of concepts. In 2014 VGGNet \cite{Simonyan_VGG_2014} introduced several architectures, with increasing depth, 11 being the shallowest and 19 the deepest, they used $3 \times 3$ conv kernels, and showed that stacking smaller kernels results in better non-linearity and achieves better accuracy. They showed the deeper, the better. The same year, GoogleNet \cite{Szegedy_googlenet_2015} was released, with 56 convolutional layers making up a 22 modular layered network, their architecture was made up of convolutional layers with $1 \times 1$, $3 \times 3$ and $5 \times 5$ kernels which they call, an Inception module. Using this architecture they could decrease the number of parameters drastically compared to former architectures. They ranked first in ImageNet challenge that year. They later revised their architecture and used two consecutive $3 \times 3$ conv layers with 128 kernels instead of the previous $5 \times 5$ layers, they also used a technique called Batch-Normalization \cite{Ioffe_BatchNorm_incepv2_2015} for reducing internal covariate shift. This technique provided improvements in several sections which is explained thoroughly in \cite{Ioffe_BatchNorm_incepv2_2015}. They achieved state of the art results in ImageNet challenge.\\ 
In 2015 prior to GoogleNet achieving the state of the art on ImageNet, \cite{He_PReLU_2015}, released their paper in which they used a ReLU variant called, Parametric RELU ( PReLU) to improve model fitting, they also created a initialization method specifically aimed at rectified nonlinearities, by which they could train deeper architectures better. Using these techniques, they could train a slightly modified version of VGGNet19 \cite{Simonyan_VGG_2014} architecture and achieve state of the art result on ImageNet. At the end of 2015, they proposed a deep architecture of 152 layers, called Residual Network (ResNet) \cite{He_ResNet_2015} which was built on top their previous findings and achieved state of the art on ImageNet previously held by themselves. In ResNet they used what they call residual blocks in which layers are let to fit a residual mapping. They also used shortcut connections to perform identity mapping. This made them capable of training deeper networks easily and gain more accuracy by going deeper without becoming more complex. In fact their model is less complex than the much shallower VGGNet \cite{Simonyan_VGG_2014} which they previously used. They investigated architectures with 1000 layers as well. Later \cite{Huang_DeepNN_StochDepth_2016} further enhanced ResNet with stochastic depth, where they used a training procedure, in which they would train a shorter network and then at test time, use a deeper architecture. Using this method they could train even deeper architectures and also achieve state of the art on CIFAR10 dataset.\\ 
Prior to the residual network, \cite{Srivastava_HighwayNets_2015} released their Long Short Term Memory (LSTM) recurrent network inspired highway networks in which they used the initialization method proposed by \cite{He_PReLU_2015} and created a special architecture that uses adaptive gating units to regulate the flow of information through the network. They created a 100 layer and also experimented with a 1K layer network and reported the easy training of such networks compared to the plain ones. Their contribution was to show that deeper architectures can be trained with Simple stochastic gradient descent.\\
In 2016 \cite{Szegedy_inceptiov4_2016} investigated the effectiveness of combining residual connections with their inceptionv3 architecture. They gave empirical evidence that training with residual connections accelerates the training of Inception networks significantly, and reported that residual Inception networks outperform similarly expensive Inception networks by a thin margin. With these variations the single-frame recognition performance on the ILSVRC 2012 classification task \cite{Russakovsky_ImageNet_2015} improves significantly. With an ensemble of three residual and one Inception-v4, they achieved 3.08 percent top-5 error on the test set of the ImageNet classification challenge. The same year, \cite{Zagoruyko_WRN_2016} ran a detailed experiment on residual nets \cite{He_ResNet_2015} and came up with a novel architecture called Wide Residual Net (WRN) where instead of a thin deep network, they increased the width of the network in favor of its depth(decreased the depth). They showed that the new architecture does not suffer from the diminishing feature reuse problem \cite{Srivastava_HighwayNets_2015} and slow training time. They report that a 16 layer wide residual network, outperforms any previous residual network architectures. They experimented with varying depth of their architecture from 10 to 40 layers and achieved state of the art result on CIFAR10/100 and SVHN. 

\subsection{Model Compression}
The computational and memory usage overhead caused by such practices, limits the expansion and applications of deep learning methods. There have been several attempts in the literature to get around such problems. One of them is model compression in which it is tried to reduce the computational overhead at inference time. It was first researched by \cite{Bucilua_Model_compression_2006}, where they tried to create a network that performs like a complex and large ensemble. In their method they used the ensemble to label unlabeled data with which they train the new neural network, thus learning the mappings learned by the ensemble and achieving similar accuracy. This idea is further worked on by \cite{Ba_DoDeepNets_nd_B_Deep_2014}. They proposed a similar concept but this time they tried to compress deep and wide networks into shallower but even wider ones. \cite{Hinton_Distillation_2015} introduced their model compression model, called Knowledge Distillation (KD), which introduces a teacher/student paradigm for transferring the knowledge from a deep complex teacher model or an ensemble of such, to less complex yet still similarly deep but fine-grained student models, where each student model can provide similar performance overall and perform better on fine-grained classes where the teacher model confuses and thus eases the training of deep networks. Inspired by \cite{Hinton_Distillation_2015}, \cite{Romero_Fitnet_2014} proposed a novel architecture to address what they referred to as not taking advantage of depth in the previous works related to Convolutional Neural Networks model compression. Previously, all works tried to compress a teacher network or an ensemble of networks into either networks of similar width and depth or into shallower and wider ones. However, they proposed a novel approach to train thin and deep networks, called FitNets, to compress wide and shallower (but still deep) networks. Their method is based on Knowledge Distillation (KD)\cite{Hinton_Distillation_2015} and extends the idea to allow for thinner and deeper student models. They introduce intermediate-level hints from the teacher hidden layers to guide the training process of the student, they showed that their model achieves the same or better accuracy than the teacher models.

\subsection{Network Pruning} 
In late 2015 \cite{Han_deep_compression_2015} released their work on model compression. They introduced “deep compression”, a three stage pipeline: pruning, trained quantization and Huffman coding, that work together to reduce the storage requirement of neural networks by 35 to 49 times without affecting their accuracy. In their method, the network is first pruned by learning only the important connections. Next, the weights are quantized to enforce weight sharing, finally, the Huffman coding is applied. After the first two steps they retrain the network to fine tune the remaining connections and the quantized centroids. Pruning, reduces the number of connections by 9 to 13 times; Quantization then reduces the number of bits that represent each connection from 32 to 5. On the ImageNet dataset, their method reduced the storage required by AlexNet by 35 times, from 240MB to 6.9MB, without loss of accuracy.

\subsection{Light weight architectures} 
In 2014 \cite{Springenberg_StrivingForSimplicity_2014} released their paper where the effectiveness of simple architectures was investigated. The authors intended to come up with a simplified architecture, not necessarily shallower, that would perform better than at the time, more complex networks. Later in 2015, they proposed different versions of their architecture and studied their characteristics, and using a 17 layer version of their architecture they achieved a result very close to state of the art on CIFAR10 with intense data-augmentation.\\
In 2016 \cite{Iandola_squeezenet_2016} released their paper in which they proposed a novel architecture called, SqueezeNet, a small CNN architecture that achieves AlexNet-level accuracy on ImageNet With 50 times fewer parameters. To our knowledge this is the first architecture that tried to be small and yet be able to achieve a good accuracy.\\
In this paper, we tried to come up with a simple architecture which exhibits the best characteristics of these works and propose a 13 layer convolutional network that achieves state of the art result on CIFAR10\footnote{\small{While preparing our paper we found out, our record was beaten by Wide Residual Net, which we then addressed in related works. We still have the state of the art record without data-augmentation as of zero padding and normalization. We also have the state of the art in terms of accuracy/parameters ratio.}}. Our network has fewer parameters (2 to 25 times less) compared to all previous deep architectures, and performs either superior to them or on par despite the huge difference in number of parameters and depth. For those architectures such as SqueezeNet/FitNet where the number of parameters is less than ours but also are deeper, our network accuracy is far superior to what can be achieved with such networks. Our architecture is also the smallest (depth wise) architecture that both has a small number of parameters compared to all leading deep architectures, and also unlike previous architectures such as SqueezeNet or FitNet, gives higher or very competitive performance against all deep architectures. 
Our model then can be compressed using deep compression techniques and be further enhanced, resulting in a very good candidate for many scenarios. 

\section{Proposed Architecture and Design intuition} \label{sec:arch}
\begin{figure*}[h]
\begin{center}
\includegraphics[width=1\linewidth]{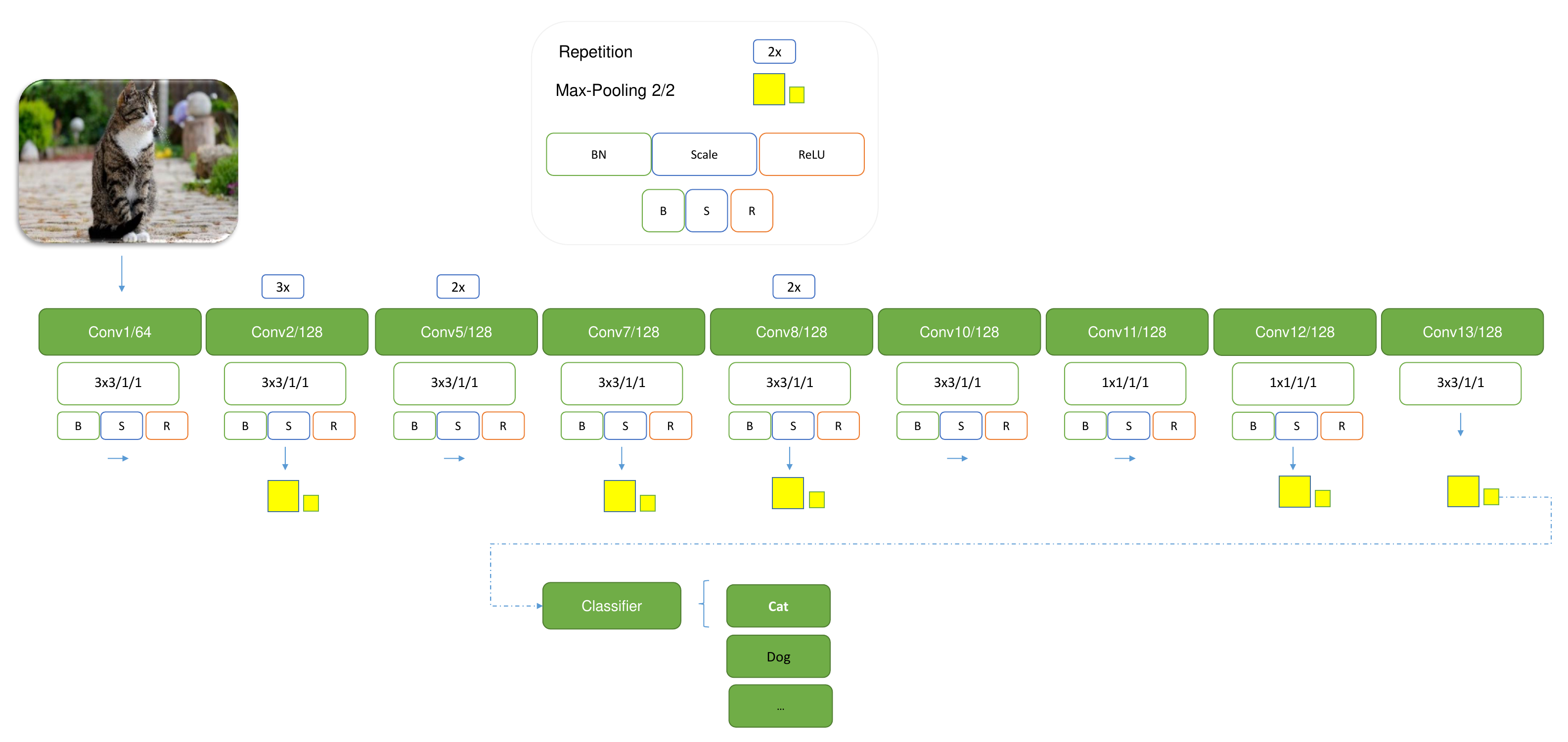}
\end{center}
 \caption{Showing the base architecture with no drop-out}
\label{fig:The_arch_figure}
\end{figure*}

We propose a simple convolutional network with 13 layers. The network employs a homogeneous design utilizing $3 \times 3$ kernels for convolutional layer and $2 \times 2$ kernels for pooling operations. Figure \ref{fig:The_arch_figure} illustrates the proposed architecture.

The only layers which do not use $3 \times 3$ kernels are 11th and 12th layers, these layers, utilize $1 \times 1$ convolutional kernels. Feature-map down-sampling is carried out using nonoverlaping $2 \times 2$ max-pooling. In order to cope with the problem of vanishing gradient and also over-fitting, we used batch-normalization with moving average fraction of 0.95 before any ReLU non-linearity. We also used weight decay as regularizer. A second version of the architecture uses dropout to cope with over-fitting. Table \ref{tab:arch_stat} shows different architectures and their statistics, among which our architecture has the lowest number of parameters and operations. The extended list is provided in the appendix.

\begin{table*}[h!]
\caption{showing different architectures statistics}
\begin{center}
\begin{tabular}{lcccccc}
Model & AlexNet & GoogleNet & ResNet152 & VGG16 & NIN & \textbf{SimpleNet} \\
\hline
Param & 60M & 7M & 60M	& 138M	& 7.6M	& \textbf{5.4M}\\
OP &7.27G & 16.04G & 11.3G & 154.7G &	11.06G & \textbf{652M}\\
Storage (MB) &217 & 40 & 230 & 512.24 & 29 & \textbf{20}\\
\hline
\end{tabular}
\end{center}
\label{tab:arch_stat}
\end{table*}

We used several principles in our work that helped us manage different issues much better and achieve desirable results. Here we present these principles with a brief explanation concerning the intuitions behind them: 

\subsubsection{Gradual Expansion and Minimum allocation}
In order to better manage the computational overhead, parameter utilization efficiency, and also network generalization power, start with a small and thin network, and then gradually expand it. Neither the depth nor the number of parameters are good indicators of how a network should perform. They are neutral factors that are only beneficial when utilized mindfully, otherwise, the design would result in an inefficient network imposing unwanted overhead. Furthermore, fewer learnable parameters also decrease the chance of overfitting and together with an enough depth it increases the network's generalization power. In order to utilize both depth and parameters more efficiently, design the architecture in a symmetric and gradual fashion, i.e. instead of creating a network with a random yet great depth, and large number of neurons per layer, start with a small and thin network then gradually add more symmetric layers. Expand the network to reach a cone shaped form. A Large degree of invariance to geometric transformations of the input can be achieved with this progressive reduction of spatial resolution compensated by a progressive increase of the richness of the representation (the number of feature maps), hence getting a conned shape, that's one of the reasons why deeper is better) \cite{Lecun_GradientBased_CNN_1998}. Therefore a deeper network with thinner layers, tends to perform better than the same network being much shallower with wider layers. It should however be noted that, very deep and very thin architectures, like their shallow and very wide counterparts are not recommended. The network needs to have proper processing and representational capacity and what this principle suggests is a method of finding the right value for depth and width of a network for this very reason. 

\subsubsection{Homogeneous Groups of Layers}
Instead of thinking in layers, think and design in group of homogeneous layers. The idea is to have several homogeneous groups of layers, each with gradually more width. The symmetric and homogeneous design, allows to easily manage the number of parameters a network will withhold and also provide better information pools for each semantic level.  
\subsection{Local Correlation Preservation}
Preserve locality information throughout the network as much as possible by avoiding $1 \times 1$ kernels in early layers. The cornerstone of CNN success lies in local correlation preservation. Avoid using $1 \times 1$ kernels or fully connected layers where locality of information matters. This includes exclusively the early layers in the network. $1 \times 1$ kernels have several desirable characteristics such as increasing networks non-linearity and feature fusion \cite{Lin_NIN_2013} which increases abstraction level, but they also ignore any local correlation in the input. Since they do not consider any neighborhood in the input and only take channels into account, they distort valuable local information. Preferably use $1 \times 1$ kernels at the end of the network or if one intends on using tricks such as bottleneck employed by GoogleNet \cite{Szegedy_googlenet_2015} and ResNet \cite{He_ResNet_2015}, use more layers with skip connections to compensate the loss in information. It is suggested to replace $1 \times 1$ kernels with $2 \times 2$ if one plans on using them other than the end of the network. Using $2 \times 2$ kernels both help to reduce the number of parameters and also to retain neighborhood information. 
\subsection{Maximum Information Utilization}
Utilize as much information as it is made available to a network by avoiding rapid down sampling especially in early layers. To increase a network's discriminative power, more information needs to be made available. This can be achieved either by a larger dataset or larger feature-maps. If larger dataset is not feasible, the existing training samples must be efficiently harnessed. Larger feature-maps especially in early layers, provide more valuable information to the network than the smaller ones. With the same depth and number of parameters, a network which utilizes bigger feature-maps achieves a higher accuracy. Therefore instead of increasing the complexity of a network by increasing its depth and number of parameters, one can leverage more performance/accuracy by simply using larger input dimensions or avoiding rapid early down-sampling. This is a good technique to keep the complexity of the network in check and improve the network performance. 
\subsection{Maximum Performance Utilization}
\begin{figure*}[h]
\begin{center}
\includegraphics[width=0.7\linewidth]{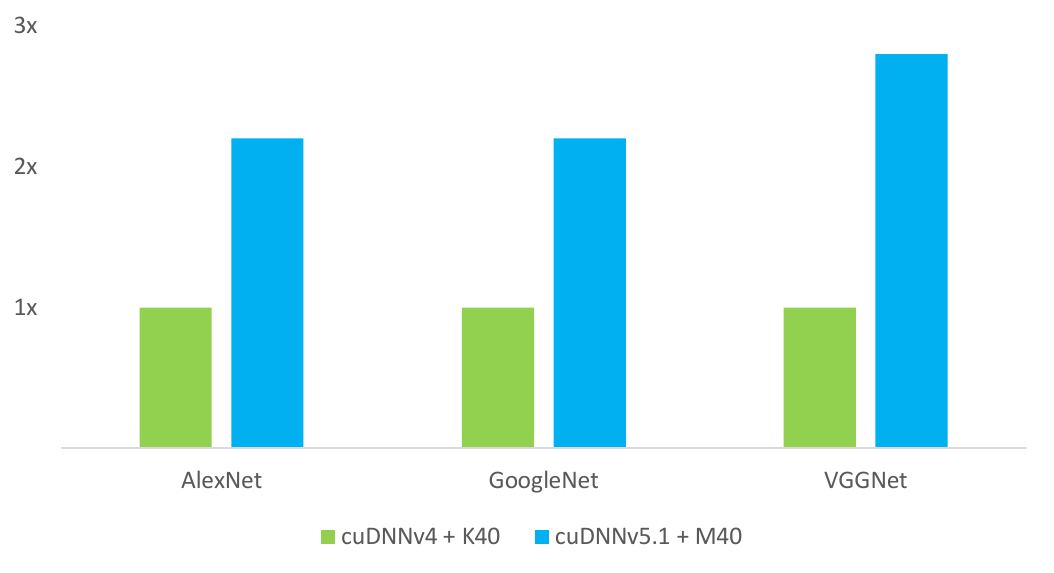}
\end{center}
 \caption{Using $3 \times 3$ kernels results in $2.7\times$ faster training when using cuDNNv$5.\times$}
\label{fig:cudnn_figure}
\end{figure*}

Use $3 \times 3$, and follow established industrial trends. For an architecture to be easily usable and widely practical, it needs to perform fast and decently. By taking into account the current improvements in underlying libraries, designing better performing and more efficient architectures are possible. Using $3 \times 3$ kernels, apart from already known benefits \cite{Simonyan_VGG_2014}, allows to achieve a substantial boost in performance when using NVIDIA's cuDNNv$5.\times$ library. A speed up of about $2.7\times $ compared to the former v4 version \footnote{https://developer.nvidia.com/cudnn-whatsnew}. This is illustrated in figure \ref{fig:cudnn_figure}. This ability to harness every amount of performance is a decisive criterion when it comes to production and industry. A fast and robust performance translates into, less time, decreased cost and ultimately a higher profit for business owners. Apart from the performance point of view, on one hand larger kernels do not provide the same efficiency per parameter as a $3 \times 3$ kernel does. It may be theorized that since larger kernels capture a larger area of neighborhood in the input, using them may help in ignoring noises and thus capturing better features, or more interesting correlations in the input because of larger receptive field and ultimately improving performance. But in fact the overhead they impose in addition to the loss in information they cause make them not an ideal choice. This makes the efficiency per parameter to decrease and causes unnecessary computational burden. Moreover, larger kernels can be replaced with a cascade of smaller ones (e.g. $3 \times 3$) which will still result in the same effective receptive field and also more nonlinearity, making them a better choice over larger kernels. 

\subsection{Rapid Prototyping}
Test the architecture with different learning policies before altering it. Most of the time, it's not the architecture that needs to be changed, rather it's the optimization policy. A badly chosen optimization policy leads to bad convergence, wasting network resources. Simple things such as learning rates and regularization methods, usually have an adverse effect if not tuned correctly. Therefore it is first suggested to use an automated optimization policy to run quick tests and when the architecture is finalized, the optimization policy is carefully tuned to maximize network performance.
\subsection{Experiment Isolation}
Conduct experiments under equal conditions. When testing a new feature, make sure only the new feature is being evaluated. For instance, when evaluating a $3 \times 3$ kernel against a $5 \times 5$ kernel, the overall network entropy must remain equal. It is usually neglected in different experiments and changes are not evaluated in isolation or better said, under an equal condition. This can lead to a wrong deduction and thus result in an inefficient design. In order to effectively assess a specific feature and its effectiveness in the architecture design, it is important to keep track of the changes, either caused by previous experiments or by the addition of the new feature itself, and take necessary action to eliminate the sources of discrepancies. 
\subsection{Minimum Entropy}
Like the previous principle, here we explain about the generalization power and why lower entropy matters. It is true that the more parameter a network withholds, the faster it can converge, and the more accuracy it can achieve, but it will over-fit more as well. A model with fewer number of parameters which provides better results or performs comparable to heavier models indicates the fact that the network has learned much better features based on which it is making its decision. In other words, by imposing more constrains on the amount of entropy a network has, we force the network to find and use much better and more robust features. This specifically manifests itself in the generalization power, since the network decisions are based on more important and more discriminative features. It can thus perform much better compared to a network with higher number of parameters which would easily overfit as well.    
\subsection{Final Regulation Stage}
While we try to formulate the best ways to achieve better accuracy in the form of rules or guidelines, they are not necessarily meant to be aggressively followed in all cases. These guidelines are meant to help achieve a good compromise between performance and the imposed overhead. Therefore start by designing according to the guidelines and then try to alter the architecture in order to get the best compromise according to your needs. In order to better tune your architecture, try not to alter or deviate a lot from multiple guidelines at once. Following a systematic procedure helps to avoid repetitive actions, and also obtain better understanding of what/which series of actions lead to specific outcomes that would normally be a hard task. Work on one aspect at a time until the desired outcome is achieved. Ultimately, it's all about the well balanced compromise between performance/imposed overhead according to one's specific needs.\\

As we have already briefly discussed in previous sections, the current trend in the community has been to start with a deep and big architecture and then use different regularization methods to cope with over-fitting. The intuition behind such a trend is that it is naturally difficult to come up with an architecture with the right number of parameters/depth that suits exactly one's data requirements. While such intuition is plausible and correct, it is not without flaws.\\
One of the issues is the fact that there are many use cases and applications for which there is not a huge dataset (such as ImageNet e.g.) available. Apart from the fact that less computation and memory overhead is always desirable for any circumstances and results in decreased costs, the majority of applications have access to medium/small sized datasets and yet they are already exploiting the benefits of deep learning and achieving either state of the art or very outstanding results. Individuals coming from this background, have two paths before them when they want to initiate a deep learning related project: 1) they either are going to design their own architecture which is difficult and time-consuming and has its own share of issues and 2) Use one of the existing heavy but very powerful architectures that have won competitions such as ImageNet or performed well on a related field of interest.\\
Using these kinds of architectures imposes a lot of overhead and users should also bear the cost of coping with the resulting over-fitting. It adversely affects training time, making it more time and resource consuming. When such architectures are used for fine-tuning, the issues caused by such deep and heavy architectures such as computational, memory and time overhead, are also imposed.\\ 
Therefore it makes more sense to have a less computationally expensive architecture which provides higher or comparable accuracy compared to the heavier counterparts. The lowered computational overhead results in a decreased time and power consumption which is a decisive factor for mobile applications. Apart from such benefits, reliance on better and more robust features is another important reason to opt for such networks.

\section{Experiments} \label{sec:exp}
We experimented on CIFAR-10/100 \cite{Krizhevsky_LearningMultipleLayers_2009}, SVHN \cite{Netzer_ReadingDigits_2011}, MNIST \cite{Lecun_GradientBased_CNN_1998} and ILSVRC 2012 classification task \cite{Russakovsky_ImageNet_2015} datasets in order to evaluate and compare our architecture against the top ranking methods and deeper models that also experimented on such datasets. We only used simple data augmentation of zero padding, and mirroring on CIFAR10/100. Other experiments on MNIST , SVHN datasets are conducted without data-augmentation. Details on data augmentation used on ImageNet dataset is explained in the respective section. In our experiments we used one configuration for all datasets and, we did not fine-tune anything except CIFAR10. We did this to see how this configuration can perform with no or slightest change in different scenarios. We used Caffe framework \cite{Jia_Caffe_2014} for training our architecture and ran our experiments on a system with an Intel Pentium G3220 CPU,14 Gigabytes of RAM, and NVIDIA GTX980. \footnote {ImageNet results are achieved using Pytorch}  
 
\subsection{CIFAR10/100}
The CIFAR10/100 \cite{Krizhevsky_LearningMultipleLayers_2009} datasets includes 60,000 color images of which 50,000 belong to the training set and 10,000 are reserved for testing (validation). These images are divided into 10 and 100 classes respectively and classification performance is evaluated using top-1 error. Table \ref{tab:cifar} shows the results achieved by different architectures.\\ 
We tried two different configurations for CIFAR10 experiment, one with no data-augmentation i.e. zero-padding and normalization, and another one using data-augmentation. We name them Arch1 and Arch2 respectively. The Arc1 achieves a new state of the art in CIFAR10 when no data-augmentation is used and the Arc2 achieves 95.51\% on CIFAR10 and 78.37\% on CIFAR100 outperforming all major architectures except the 36m variant of WRN. More results are provided in the appendix. 

\begin{table}[H]
\centering
\caption{Top CIFAR10/100 results.}
\label{tab:cifar}
\begin{tabular}{lccc}
\textbf{Method}  & \textbf{\#Params}   & \textbf{CIFAR10}       & \textbf{CIFAR100}      \\ \hline
VGGNet(16L) \cite{Sergey_CIFAR10_OnTorch_2015}/Enhanced  & 138m   & 91.4 / 92.45  & -             \\ 
ResNet-110L / 1202L \cite{He_ResNet_2015} *   & 1.7/10.2m & 93.57 / 92.07 & 74.84/72.18 \\ 
SD-110L / 1202L \cite{Huang_DeepNN_StochDepth_2016} & 1.7/10.2m & 94.77 / 95.09 & 75.42 / -     \\ 
WRN-(16/8)/(28/10) \cite{Zagoruyko_WRN_2016} & 11/36m    & 95.19 / 95.83 & 77.11/79.5  \\ 
Highway Network \cite{Srivastava_HighwayNets_2015}  & N/A   & 92.40  & 67.76    \\ 
FitNet \cite{Romero_Fitnet_2014}   & 1M  & 91.61  & 64.96   \\ 
FMP* (1 tests) \cite{Graham_FractionalMaxpooling_2014}     & 12M    & 95.50   & 73.61         \\ 
Max-out(k=2) \cite{Goodfellow_MaxoutNetwork_2013}    & 6M  & 90.62   & 65.46         \\ 
Network in Network \cite{Lin_NIN_2013}   & 1M   & 91.19    & 64.32   \\ 
DSN \cite{Lee_DeeplySupervisedNet_2015}   & 1M   & 92.03   & 65.43         \\ 
Max-out NIN \cite{JiaRen_BatchNormMaxoutNIN_2015}    & -     & 93.25         & 71.14         \\ 
LSUV \cite{Mishkin_AllYouNeedIsGoodInit_2016}    & N/A    & 94.16    & N/A     \\ 
SimpleNet-Arch 1$*$       & 5.48M      & \textbf{94.75}   & -         \\ 
SimpleNet-Arch 2 $\dagger$     & 5.48M      & \textbf{95.51}   & \textbf{78.37}         \\ \hline
\end{tabular}
\end{table}

*Note that the Fractional Max Pooling \cite{Graham_FractionalMaxpooling_2014} uses a deeper architecture and also uses extreme data augmentation. $*$  means No zero-padding or normalization with dropout and $\dagger$ means Standard data-augmentation- with dropout. To our knowledge, our architecture has the state of the art result, without the aforementioned data-augmentations.

\subsection{ MNIST}
The MNIST dataset \cite{Lecun_GradientBased_CNN_1998} consists of 70,000 28x28 grayscale images of handwritten digits 0 to 9, of which 60,000 are used for training and 10,000 are used for testing. We didn't use any data augmentation on this dataset, and yet scored second to the state-of-the-art without data-augmentation and fine-tuning. We also slimmed our architecture to have only 300K parameters and achieved 99.72\% accuracy beating all previous larger and heavier architectures .Table \ref{tab:MNIST} shows the current state of the art results for MNIST.


\begin{table}[H]
\centering
\caption{Top MNIST results}
\label{tab:MNIST}
\begin{tabular}{lc}
\textbf{Method} & \textbf{Error rate} \\ \hline
DropConnect\cite{Wan_Regularization_Using_DropConnect_2013}** & 0.21\% \\
Multi-column DNN for Image Classiﬁcation\cite{Ciregan_Multi_column_dnn_img_cls_2012}**  & 0.23\% \\
APAC\cite{Sato_APAC_2015}** & 0.23\%\\
Generalizing Pooling Functions in CNN\cite{Lee_CNN_Mixed_gated_2016}** & 0.29\%\\
Fractional Max-Pooling\cite{Graham_FractionalMaxpooling_2014}** & 0.32\%\\
Batch-normalized Max-out NIN \cite{JiaRen_BatchNormMaxoutNIN_2015}   & 0.24\%     \\ 
Max-out network (k=2)  \cite{Goodfellow_MaxoutNetwork_2013}          & 0.45\%     \\ 
Network In Network \cite{Lin_NIN_2013}                              & 0.45\%     \\ 
Deeply Supervised Network  \cite{Lee_DeeplySupervisedNet_2015}       & 0.39\%     \\ 
RCNN-96 \cite{Liang_RecurrentCNN_2015}                              & 0.31\%     \\ 
\textbf{SimpleNet *}                               & \textbf{0.25\% }    \\ \hline
\end{tabular}
\end{table}
*Note that we didn't intend on achieving the state of the art performance here as we are using a single optimization policy without fine-tuning hyper parameters or data-augmentation for a specific task, and still we nearly achieved state-of-the-art on MNIST. **Results achieved using an ensemble or extreme data-augmentation

\subsection{SVHN}
The SVHN dataset \cite{Netzer_ReadingDigits_2011} is a real-world image dataset, obtained from house numbers in Google Street View images. It consists of 630,420 32x32 color images of which 73,257 images are used for training, 26,032 images are used for testing and the other 531,131 images are used for extra training. Like \cite{Huang_DeepNN_StochDepth_2016, Goodfellow_MaxoutNetwork_2013, Lin_NIN_2013} we only used the training and testing sets for our experiments and didn't use any data-augmentation. We also used the slimmed version with 300K parameters and obtained a very good test error of 2.37\%. Table \ref{tab:SVHN} shows the current state of the art results for SVHN.


\begin{table}[h!]
\caption{Top SVHN results.}\label{tab:SVHN}
\begin{center}
\begin{tabular}{lc}
\textbf{Method} & \textbf{Error rate} \\ \hline
Network in Network\cite{Lin_NIN_2013}    &	2.35 \\
Deeply Supervised Net\cite{Lee_DeeplySupervisedNet_2015}	& 1.92 \\
ResNet\cite{He_ResNet_2015} (reported by \cite{Huang_DeepNN_StochDepth_2016} (2016)) &	2.01 \\
ResNet with Stochastic Depth\cite{Huang_DeepNN_StochDepth_2016}	&1.75\\
Wide ResNet\cite{Zagoruyko_WRN_2016}  &	1.64\\
\textbf{SimpleNet}	& \textbf{1.79}\\\hline
\end{tabular}
\end{center}
\end{table}

\subsection{Extended test}
Some architectures can't scale well when their processing capacity decreases. This shows the design is not robust enough to efficiently use its processing capacity. We tried a slimmed version of our architecture which has only 300K parameters to see how it performs and whether it's still efficient. The network also does not use any dropout. Table \ref{tab:Slimmed} shows the results for our architecture with only 300K parameters in comparison to other deeper and heavier architectures with 2 to 20 times more parameters. 



\begin{table}[H]
	\begin{center}
\caption{Slimmed version results on CIFAR10/100 datasets.}\label{tab:Slimmed}
		\begin{tabular}{lccc}
Model & Param & CIFAR10 &CIFAR100 \\ \hline
\textbf{SimpleNet} & \textbf{310K - 460K} &\textbf{91.98 - 92.33}& \textbf{64.68 - 66.82} \\
Maxout \cite{Goodfellow_MaxoutNetwork_2013} & 6M & 90.62 & 65.46 \\
DSN \cite{Lee_DeeplySupervisedNet_2015} & 1M &  92.03 & 65.43 \\
ALLCNN \cite{Springenberg_StrivingForSimplicity_2014} & 1.3M& 92.75& 66.29 \\
dasNet \cite{stollenga_dasnet_2014} & 6M & 90.78 & 66.22 \\
ResNet \cite{He_ResNet_2015} {\tiny(Depth32, tested by us)} & 475K &  91.6 & 67.37 \\
WRN \cite{Zagoruyko_WRN_2016}  & 600K &  93.15 & 69.11 \\
NIN \cite{Lin_NIN_2013}  & 1M	&91.19 &---\\ \hline
\end{tabular}
	\end{center}
	
\end{table}

\subsection{ImageNet Results}
ImageNet\cite{Russakovsky_ImageNet_2015} dataset includes images of 1000 classes and is split into three sets: 1.2M training images, 50K validation images, and 100K testing images. The classification performance is evaluated using two measures: the top-1 and top-5 errors.\\
For this experiment, the architecture needs minor changes in order to efficiently accommodate the much larger input size of 224x224. Since the input is much larger now, it needs to be downsampled for decreased computation overhead otherwise it will incur a huge processing and memory overhead during training. 
Rapid downsampling lowers the overhead at the expense of hurting the learning process, thus in order to maximize the information utilization, while maintaining a low overhead, we use strides of 2 on layers 1,2 and 4. We then remove two pooling layers to further accentuate this effect, all other aspects of the network remain intact. More information can be found in the appendix.\\ 
For training, we used the TensorFlow RMSPropOptimizer\footnote{The Pytorch implementation by \cite{rw2019timm}} with the momentum value of $0.9$ and $\epsilon=0.001$ and an initial learning rate of $0.0195$. We used a learning rate decay rate of $0.0981$ every epoch. Furthermore, we didn't use any dropouts until the network neared its saturation point, at which point, dropout values of $0.02,0.05,0.05$ were used for $10th$ to $12th$ layers respectively. The values used for l2 weight decay range from $1e-5$ to $3e-5$ depending on the architecture variant.  
We followed Inception \cite{Szegedy_inceptiov4_2016} for image preprocessing. Table \ref{tab:imagenet_experiment} shows a comparison between our architecture and major heavier and more complex architectures. \footnote{We used the average weights for some models and fine-tuned them to quickly train our model.}

\begin{table}[h!]
\caption{SimpleNet outperforms much deeper and larger architectures on the ImageNet dataset.}\label{tab:imagenet_experiment}
\begin{center}
\begin{tabular}{lccc}
\textbf{Method} & \textbf{Param} &\textbf{Top1 Accuracy} &\textbf{Top5 Accuracy}\\ \hline
AlexNet \cite{AlexKrizhevsky_imgnet_2012}   & 60M &	57.2 & 80.3 \\
SqeezeNet \cite{Iandola_squeezenet_2016}   & 1.2M &	58.18 & 80.62 \\
VGGNet16 \cite{Simonyan_VGG_2014}	& 138M & 71.59  & 90.38 \\
VGGNet16\_BN \cite{Simonyan_VGG_2014}	& 138M & 73.36 & 91.52 \\
VGGNet19 \cite{Simonyan_VGG_2014}	& 143M & 72.38 & 90.88 \\
VGGNet19\_BN \cite{Simonyan_VGG_2014}	& 143M & 74.22 & 91.84 \\
GoogleNet \cite{Szegedy_googlenet_2015}  & 6.6M &	69.78 & 89.53 \\
WResNet18 \cite{Zagoruyko_WRN_2016}  & 11.7M & 69.60 & 89.07\\
ResNet18 \cite{He_ResNet_2015}	& 11.7M &  69.76 & 89.08\\
ResNet34 \cite{He_ResNet_2015}	& 21.8M &  73.31 & 91.42\\
SimpleNet\_small\_050	& \textbf{1.5M} & \textbf{61.67} & \textbf{83.49}\\
SimpleNet\_small\_075	& \textbf{3.2M} & \textbf{68.51} & \textbf{88.15}\\
SimpleNet\_5m	& \textbf{5.7M} & \textbf{72.03} & \textbf{90.32}\\
SimpleNet\_9m & \textbf{9.5M} &	\textbf{74.23} & \textbf{91.75}\\\hline
\end{tabular}
\end{center}
\end{table}

* Note that for SimpleNet, the m2 variants are reported here\footnote{Ongoing training results} and for other given variants, the new and improved results from the official Pytorch\cite{paszke2019pytorch} repository are reported.\footnote{ \url{https://github.com/pytorch/vision/blob/bac678c8897cb8ebbca1d3877350288993b6ca69/torchvision/models/} }

\section{Conclusion} \label{sec:conclusion}
In this paper, we proposed a simple convolution architecture that takes advantage of the simplicity in its design and outperforms deeper and more complex architectures in spite of having considerably fewer parameters and operations. We showed that a good design should be able to efficiently use its processing capacity and showed that our slimmed version of the architecture with a much fewer number of parameters (300K) also outperforms deeper and or heavier architectures. Intentionally limiting ourselves to a few layers and basic elements for designing an architecture allowed us to overlook the unnecessary details and concentrate on the critical aspects of the architecture, keeping the computation in check and achieving high efficiency.\\
We tried to show the importance of simplicity and optimization using our experiments and also encourage more researchers to study the vast design space of convolutional neural networks in an effort to find more and better guidelines to make or propose better-performing architectures with much less overhead. This will hopefully greatly help to expand deep learning related methods and applications, making them more viable in more situations.\\ 
Due to a lack of good hardware, we had to contend with a few configurations. We are still continuing our tests and would like to extend our work by experimenting with new applications and design choices especially using the latest achievements about deep architectures in the literature.

\section{Acknowledgement} \label{sec:ackn}
We would like to express our deep gratitude to Dr. Ali Diba the CTO of Sensifai for his great help and cooperation in this work. We also would like to express our great appreciation to Dr. Hamed Pirsiavash for his insightful comments and constructive suggestions. We would also like to thank Dr. Reza Saadati, and Dr. Javad Vahidi for their valuable help in the early stages of the work.

\bibliography{lets_keepit_simple}
\bibliographystyle{lets_keepit_simple}

\newpage
\renewcommand{\appendixname}{Annex}
\appendix
\section{Appendix}
\subsection{ImageNet Experiment}
For the ImageNet experiment, we evaluated two variants of our model, namely m1 and m2. The m1 variant incorporates strides of 2 in the first three layers, while the m2 variant utilizes strides of 2 in layers 1, 2, and 4. Delaying the downsampling by just 1 layer in the m2 variant, allows it to better utilize the information from the larger featuremaps and thus achieve a higher accuracy as it better utilizes the slightly more information at its disposal.\\
Table \ref{tab:imagenet_ext_appendix} demonstrates the difference between the two variants in terms of runtime performance and the accuracy achieved. Table \ref{tab:imagenet_ext_appendix2} shows an extended version of such comparison between SimpleNet variants and a few larger and more complex architectures.     

\begin{table}[h!]
\caption{SimpleNet variants(m1 vs m2) running on GTX1080 and Pytorch1.11. The m1 variant is faster while the m2 variant achieves higher accuracy. }\label{tab:imagenet_ext_appendix}
\begin{center}
\begin{tabular}{lccc}
\textbf{Method}          &\textbf{Samples per sec}   &\textbf{Top1 accuracy}       &\textbf{Top5 accuracy}\\ \hline
SimpleNetv1\_small\_05	      &5234 / 4466        &61.12 / 61.67       &82.99 / 83.49  \\ 
SimpleNetv1\_small\_075      &2893 / 2478        &67.78 / 68.51       &87.72 / 88.15  \\ 
SimpleNetv1\_5m              &2105 /1754         &71.55 / 72.03       &89.94 / 90.32  \\
SimpleNetv1\_9m	            &1048 / 900         &73.79 / 74.23        &91.49 / 91.75   \\ 
\end{tabular}
\end{center}
\end{table}

\begin{table}[h!]
    \caption{SimpleNet variants running on GTX1080 and Pytorch1.11. SimpleNet variants provide a good ratio between performance and accuracy. }\label{tab:imagenet_ext_appendix2}
    \begin{center}
    \begin{tabular}{lcccc}
    \textbf{Method}    &\textbf{Samples per sec}   &\textbf{Param}   &\textbf{Top1}   &\textbf{Top5}\\ \hline
    \textbf{SimpleNetv1\_small\_m1\_05}   & \textbf{5234.21}    & \textbf{1.51}    & \textbf{61.12}     & \textbf{82.99} \\
    \textbf{SimpleNetv1\_small\_m2\_05}   & \textbf{4466.09}    & \textbf{1.51}    & \textbf{61.67}     & \textbf{83.49}  \\
    \textbf{SimpleNetv1\_small\_m1\_075}   & \textbf{2893.91}    & \textbf{3.29}    & \textbf{67.784}     & \textbf{87.718} \\
    \textbf{SimpleNetv1\_small\_m2\_075}   & \textbf{2478.41}    & \textbf{3.29}    & \textbf{68.506}     & \textbf{88.15}  \\
    \textbf{SimpleNetv1\_5m\_m1}           & \textbf{2105.06}    & \textbf{5.75}    & \textbf{71.548}     & \textbf{89.94}  \\
    \textbf{SimpleNetv1\_5m\_m2}           & \textbf{1754.25}    & \textbf{5.75}    & \textbf{72.03}      & \textbf{90.324} \\
    ResNet18 \cite{He_ResNet_2015}                               & 1750.38    & 11.69            & 69.76              & 89.08  \\
    \textbf{SimpleNetv1\_9m\_m1}     & \textbf{1048.91} &\textbf{9.51}     & \textbf{73.792}     & \textbf{91.486}  \\
    ResNet34 \cite{He_ResNet_2015}                               & 1030.4     & 21.8             & 73.31                & 91.42  \\
    \textbf{SimpleNetv1\_9m\_m2}   & \textbf{900.45}    & \textbf{9.51}    & \textbf{74.23}      & \textbf{91.748}  \\
    VGGNet13 \cite{Simonyan_VGG_2014}                                   & 363.69     & 133.05           & 69.928              & 89.246  \\
    VGGNet13\_bn\ \cite{Simonyan_VGG_2014}                               & 315.85     & 133.05           & 71.586              & 90.374  \\
    VGGNet16 \cite{Simonyan_VGG_2014}                                  & 302.84     & 138.36           & 71.592               & 90.382  \\
    VGGNet16\_bn \cite{Simonyan_VGG_2014}                              & 265.99     & 138.37           & 73.360               & 91.516  \\
    VGGNet19 \cite{Simonyan_VGG_2014}                                  & 259.82     & 143.67           & 72.376              & 90.876   \\
    VGG19\_bn \cite{Simonyan_VGG_2014}                              & 229.77     & 143.68           & 74.218              & 91.842  \\
    \end{tabular}
    \end{center}
    \end{table}
    
\subsection{Extended Results}
In this section, the extended results pertaining to CIFAR10, CIFAR100 are provided in tables \ref{tab:cifar10_appndx}, \ref{tab:cifar100_appndx} and respectively.\\ 

\begin{table}[H]
\caption{CIFAR10 extended results}\label{tab:cifar10_appndx}
\begin{center}
\begin{tabular}{lcc}
\textbf{Method} & \textbf{Accuracy} & \textbf{\#Params}\\ \hline
VGGNet(16L)\cite{Sergey_CIFAR10_OnTorch_2015}  & 91.4  & 138m\\
VGGNET(Enhanced-16L)\cite{Sergey_CIFAR10_OnTorch_2015}* & 92.45	& 138m\\
ResNet-110\cite{He_ResNet_2015}* & 93.57 & 1.7m\\
ResNet-1202\cite{He_ResNet_2015} & 92.07 & 10.2m\\
Stochastic depth-110L\cite{Huang_DeepNN_StochDepth_2016} & 94.77 & 1.7m \\
Stochastic depth-1202L\cite{Huang_DeepNN_StochDepth_2016} & 95.09 & 10.2m \\
Wide Residual Net\cite{Zagoruyko_WRN_2016} & 95.19 & 11m \\
Wide Residual Net\cite{Zagoruyko_WRN_2016} & 95.83 & 36m \\
Highway Network\cite{Srivastava_HighwayNets_2015} & 92.40 & - \\
FitNet\cite{Romero_Fitnet_2014} & 91.61 & 1M \\
SqueezNet\cite{Iandola_squeezenet_2016}-(tested by us) & 79.58 & 1.3M \\
ALLCNN\cite{Springenberg_StrivingForSimplicity_2014} & 92.75 & - \\
Fractional Max-pooling* (1 tests)\cite{Graham_FractionalMaxpooling_2014} & 95.50 & 12M \\
Max-out(k=2)\cite{Goodfellow_MaxoutNetwork_2013} & 90.62 & 6M \\
Network in Network\cite{Lin_NIN_2013} & 91.19 & 1M \\
Deeply Supervised Network\cite{Lee_DeeplySupervisedNet_2015} & 92.03 & 1M \\
Batch normalized Max-out NIN\cite{JiaRen_BatchNormMaxoutNIN_2015} & 93.25 & - \\
All you need is a good init (LSUV)\cite{Mishkin_AllYouNeedIsGoodInit_2016} & 94.16 & - \\
Generalizing Pooling Functions in CNN\cite{Lee_CNN_Mixed_gated_2016} & 93.95 & - \\
Spatially-Sparse CNNs\cite{Graham_Spatilly_sparse_CNN_2014} & 93.72 & - \\
\small{Scalable Bayesian Optimization Using DNN}\cite{Snoek_ScalableBayesianoptemiz_2015} & 93.63 & - \\
Recurrent CNN for Object Recognition\cite{Liang_RecurrentCNN_2015} & 92.91 & - \\
RCNN-160\cite{Liang_RecurrentCNN_2015} & 92.91 & - \\
SimpleNet-Arch1 & 94.75 & 5.4m \\
SimpleNet-Arch1 using data augmentation & 95.51 & 5.4m \\ \hline
\end{tabular}
\end{center}
\end{table}

\begin{table}[H]
\caption{CIFAR100 extended results}\label{tab:cifar100_appndx}
\begin{center}
\begin{tabular}{lc}
\textbf{Method} & \textbf{Accuracy}\\ \hline
GoogleNet with ELU\cite{Clevert_Fast_n_accurat_ELU_2015}* & 75.72 \\
Spatially-sparse CNNs\cite{Graham_Spatilly_sparse_CNN_2014} & 75.7 \\
Fractional Max-Pooling(12M) \cite{Graham_FractionalMaxpooling_2014} & 73.61 \\
\small{Scalable Bayesian Optimization Using DNNs}\cite{Snoek_ScalableBayesianoptemiz_2015} & 72.60 \\
All you need is a good init\cite{Mishkin_AllYouNeedIsGoodInit_2016} & 72.34 \\
Batch-normalized Max-out NIN(k=5)\cite{JiaRen_BatchNormMaxoutNIN_2015} & 71.14 \\
Network in Network\cite{Lin_NIN_2013} & 64.32 \\
Deeply Supervised Network\cite{Lee_DeeplySupervisedNet_2015} & 65.43 \\
ResNet-110L\cite{He_ResNet_2015} & 74.84 \\
ResNet-1202L\cite{He_ResNet_2015} & 72.18 \\
WRN\cite{Zagoruyko_WRN_2016} & 77.11/79.5 \\
Highway\cite{Srivastava_HighwayNets_2015} & 67.76 \\
FitNet\cite{Romero_Fitnet_2014} & 64.96 \\
SimpleNet & 78.37 \\ \hline
\end{tabular}
\end{center}
\end{table}

*Achieved using several data-augmentation tricks

\begin{table}[h!]
\caption{Flops and Parameter Comparison of Models trained on ImageNet}\label{tab:Flops_appndx}
\begin{center}
\begin{tabular}{lccccccc}
\textbf{Model} & \textbf{MACC} & \textbf{COMP} & \textbf{ADD} & \textbf{DIV} & \textbf{Activations} & \textbf{Params} & \textbf{SIZE(MB)}\\ \hline
SimpleNet & 1.9G & 1.82M & 1.5M & 1.5M & 6.38M & 6.4M & 24.4 \\
SqueezeNet & 861.34M & 9.67M & 226K & 1.51M & 12.58M & 1.25M & 4.7  \\
Inception v4* & 12.27G & 21.87M & 53.42M & 15.09M & 72.56M & 42.71M & 163 \\
Inception v3* & 5.72G & 16.53M & 25.94M & 8.97M & 41.33M & 23.83M & 91 \\
Incep-Resv2* & 13.18G & 31.57M & 38.81M & 25.06M & 117.8M & 55.97M & 214\\
ResNet-152 & 11.3G & 22.33M & 35.27M & 22.03M & 100.11M & 60.19M & 230 \\ 
ResNet-50 & 3.87G & 10.89M & 16.21M & 10.59M & 46.72M & 25.56M & 97.70 \\ 
AlexNet & 7.27G & 17.69M & 4.78M & 9.55M & 20.81M & 60.97M & 217.00 \\ 
GoogleNet & 16.04G & 161.07M & 8.83M & 16.64M & 102.19M & 7M & 40 \\ 
NIN & 11.06G & 28.93M & 380K & 20K & 38.79M & 7.6M & 29 \\ 
VGG16 & 154.7G & 196.85M & 10K & 10K & 288.03M & 138.36M & 512.2 \\ \hline
\end{tabular}
\end{center}
\end{table}

*Inception v3, v4 did not have any Caffe model, so we reported their size related information from MXNet  and Tensorflow  respectively. Inception-ResNet-V2 would take 60 days of training with 2 Titan X to achieve the reported accuracy.
Statistics are obtained using \url{http://dgschwend.github.io/netscope}

\subsection{Generalization Samples}
In order to see how well the model generalizes, and whether it was able to develop robust features, we tried some images that the network has never faced and used them with a model trained on CIFAR10 dataset. As the results show, the network classifies them correctly despite the fact that they are very different from the images used for training. These visualizations are done using Deep Visualization Toolbox  by \cite{Yosinski_Understanding_deepvis_2015} and early un-augmented version of SimpleNet.

\begin{figure}[h!]
  \centering
\begin{subfigure}[b]{0.20\linewidth}
  \includegraphics[width=\linewidth]{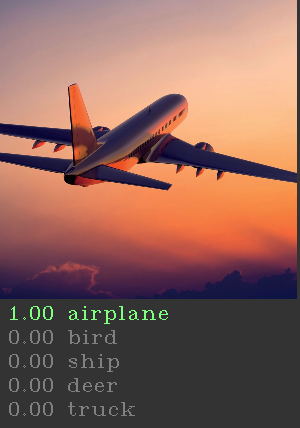}
\end{subfigure}
\begin{subfigure}[b]{0.20\linewidth}
  \includegraphics[width=\linewidth]{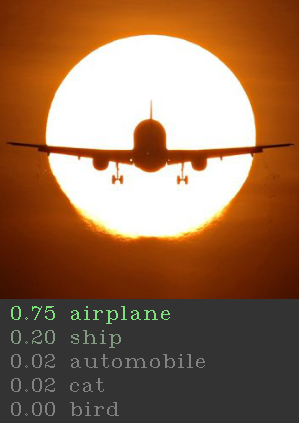}
\end{subfigure}
\begin{subfigure}[b]{0.20\linewidth}
  \includegraphics[width=\linewidth]{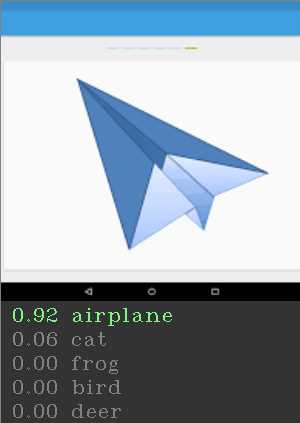}
\end{subfigure}

\begin{subfigure}[b]{0.20\linewidth}
  \includegraphics[width=\linewidth]{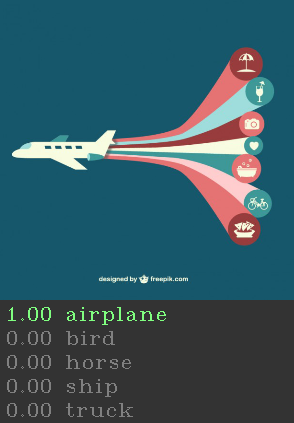}
\end{subfigure}
\begin{subfigure}[b]{0.20\linewidth}
  \includegraphics[width=\linewidth]{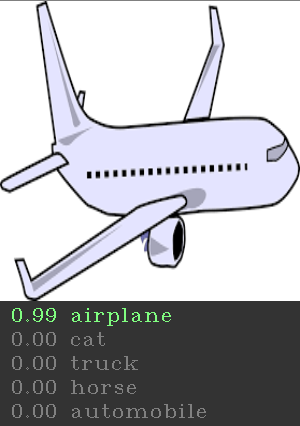}
\end{subfigure}
\begin{subfigure}[b]{0.20\linewidth}
  \includegraphics[width=\linewidth]{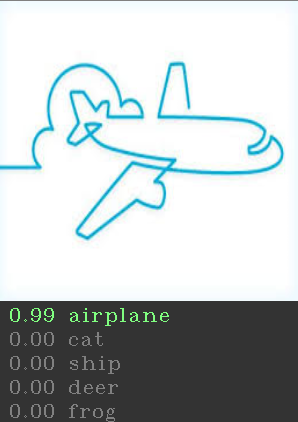}
\end{subfigure}
\caption{Some airplanes pictures with completely different appearances that the network classifies very well.}
\label{fig:fig_airplanes}
\end{figure}

\begin{figure}[h!]
  \centering
\begin{subfigure}[b]{0.20\linewidth}
  \includegraphics[width=\linewidth]{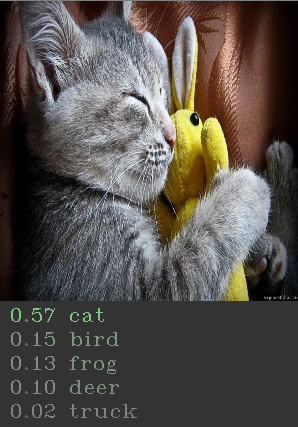}
\end{subfigure}
\begin{subfigure}[b]{0.20\linewidth}
  \includegraphics[width=\linewidth]{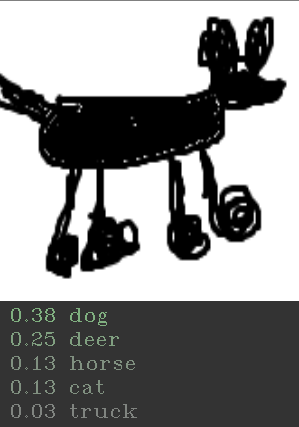}
\end{subfigure}
\begin{subfigure}[b]{0.20\linewidth}
  \includegraphics[width=\linewidth]{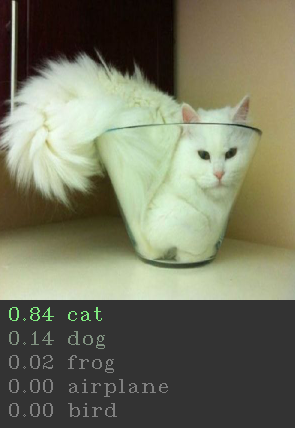}
\end{subfigure}

\begin{subfigure}[b]{0.20\linewidth}
  \includegraphics[width=\linewidth]{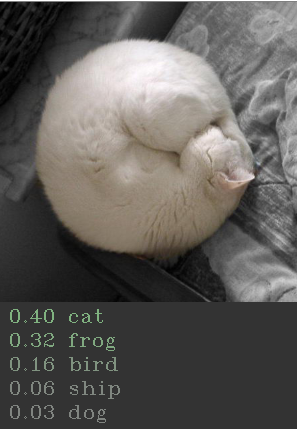}
\end{subfigure}
\begin{subfigure}[b]{0.20\linewidth}
  \includegraphics[width=\linewidth]{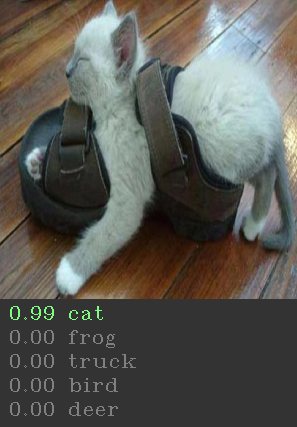}
\end{subfigure}
\begin{subfigure}[b]{0.20\linewidth}
  \includegraphics[width=\linewidth]{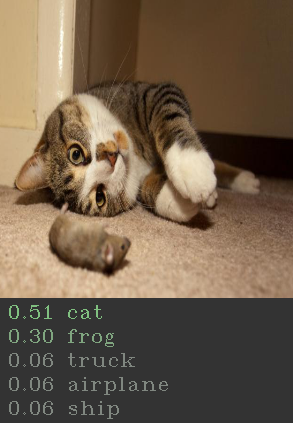}
\end{subfigure}
\caption{showing some cat images with a lot of deformations and also a drawing of animal.}
\label{fig:fig_cats}
\end{figure}
 
\begin{figure}[h!]
  \centering
\begin{subfigure}[b]{0.31\linewidth}
  \includegraphics[width=\linewidth]{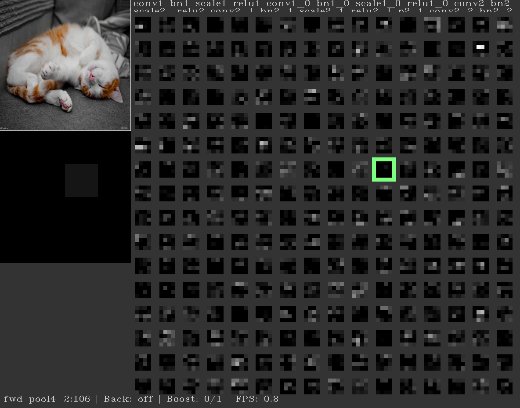}
\end{subfigure}
\begin{subfigure}[b]{0.31\linewidth}
  \includegraphics[width=\linewidth]{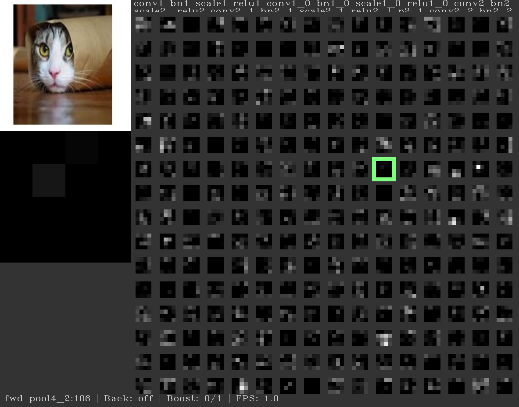}
\end{subfigure}
\begin{subfigure}[b]{0.31\linewidth}
  \includegraphics[width=\linewidth]{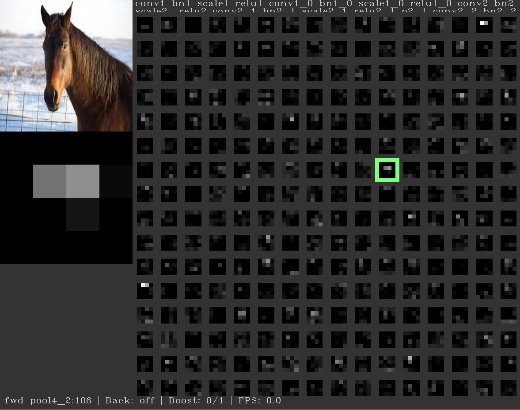}
\end{subfigure}

\begin{subfigure}[b]{0.31\linewidth}
  \includegraphics[width=\linewidth]{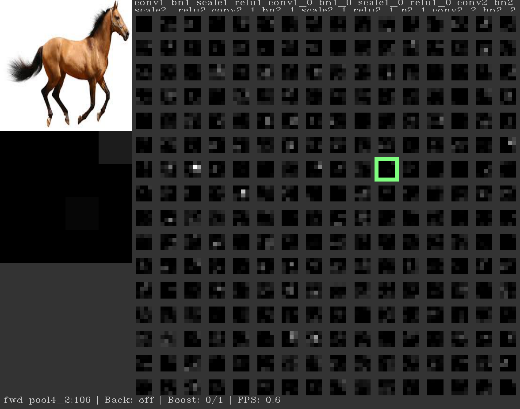}
\end{subfigure}
\begin{subfigure}[b]{0.31\linewidth}
  \includegraphics[width=\linewidth]{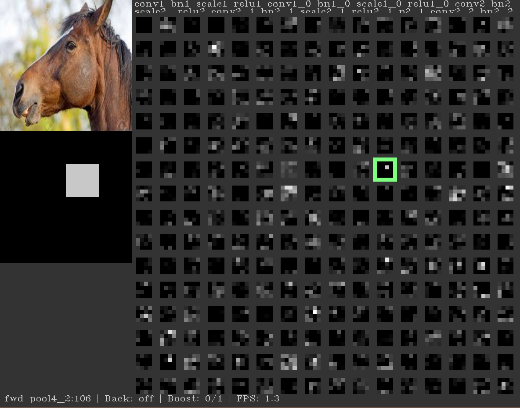}
\end{subfigure}
\begin{subfigure}[b]{0.31\linewidth}
  \includegraphics[width=\linewidth]{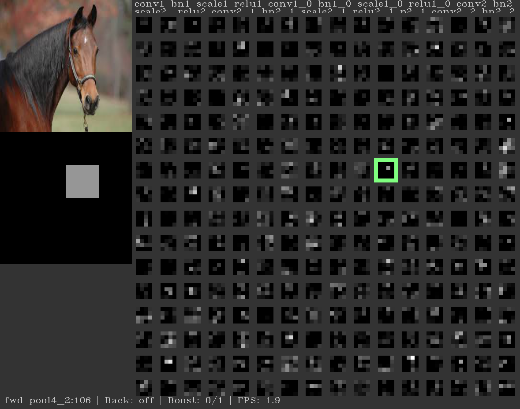}
\end{subfigure}

\begin{subfigure}[b]{0.93\linewidth}
  \includegraphics[width=\linewidth]{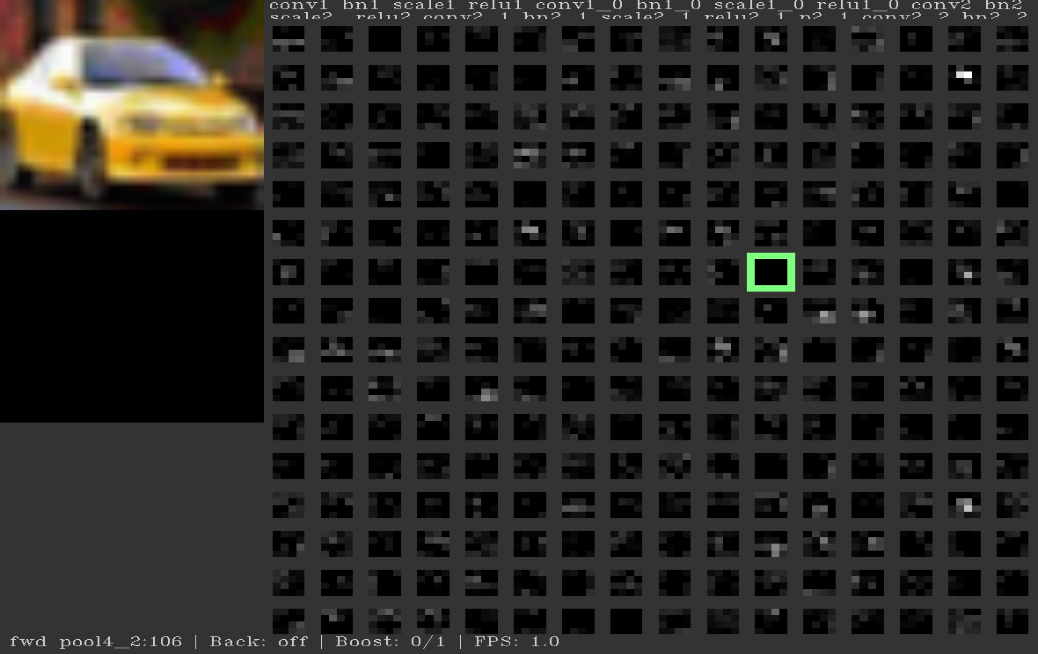}
\end{subfigure}
\caption{Showing a head detector the network has learned, which responds when facing a head in images of animals it has never seen. The same detector does not activate when confronted with an image of a car!}
\label{fig:horse_car}
\end{figure}

An interesting point in the figure \ref{fig:fig_cats} lies in the black dog/cat like drawing and the interesting predictions the network does on the strange drawing we drew! We intentionally drew a figure that does look like several categories inside CIFAR10 dataset, and thus wanted to test how it looks like to the network and whether the network uses sensible features to distinguish between each class. Interestingly the network tries its best and classifies the image according to the prominent features it finds in the picture. The similarity to some animals present in the dataset is manifested in the first four predictions and then a truck at the end denotes the circular shape of the animal's legs might have been used as an indication of the existence of the truck! Suggesting the network is trying to use prominent features to identify each class rather than some random features. Investigating the internals of the network also shows, such predictions are because of a well developed feature combinations, by which the network performs its deduction. Figure \ref{fig:horse_car} shows the network has developed a proper feature to distinguish the head/shoulder in the input, and a possible deciding factor by which to distinguish between animals and non animals. As it can be seen from the samples, while the results are very encouraging and in high confidence, they are still far from prefect. This observation may suggest 3 possible reasoning: 1) The network does not have the capability needed to perfectly deduce as we expect. 2) More data is needed for the network to develop better features, a small dataset such as CIFAR10 with no data augmentation is not simply enough to provide such capability we expect. 3) The current optimization process that we employ to train our deep architectures is insufficient and or incapable of providing such capability easily or at all. Apart from the current imperfections, results show that even a simple architecture, when properly devised, can perform decently.        

\end{document}